\pdfoutput=1
\documentclass[11pt]{article}

\usepackage{acl}

\usepackage{times}
\usepackage{microtype}
\usepackage{enumitem}
\usepackage{url}
\usepackage{graphicx}
\usepackage{tikz-dependency}
\usepackage{booktabs}
\usepackage{adjustbox}
\usepackage{color}
\usepackage{tipa}
\usepackage{comment}

\title{LLMs for POS Tagging in LRLs? \\Not there yet}
\title{Comparing LLM prompting with Cross-lingual transfer performance on Indigenous and Low-resource Brazilian Languages}

 \author{David Ifeoluwa Adelani \\  University College London, UK \\ \texttt{d.adelani@ucl.ac.uk} \And  
A. Seza Doğruöz \\  Universiteit Gent, LT3, IDLab, Gent, Belgium \\ \texttt{as.dogruoz@ugent.be} \AND
André Coneglian \\  Federal University of Minas Gerais, Brazil \\ \texttt{coneglia@ufmg.br} \And
Atul Kr. Ojha \\ Insight SFI Research Centre for Data\\ Analytics, University of Galway, Ireland \\ \texttt{atulkumar.ojha@insight-centre.org}}
\begin{document}
\maketitle
\begin{abstract}
 Large Language Models are transforming NLP for a variety of tasks. However, how LLMs perform NLP tasks for low-resource languages (LRLs) is less explored. In line with the goals of the AmericasNLP workshop, we focus on 12 LRLs from Brazil, 2 LRLs from Africa and 2 high-resource languages (HRLs) (e.g., English and Brazilian Portuguese). Our results indicate that the LLMs perform worse for the part of speech (POS) labeling of LRLs in comparison to HRLs. We explain the reasons behind this failure and provide an error analysis through examples observed in our data set.  
\end{abstract}

\section{Introduction}
\label{intro}
%We are currently witnessing major transformations in language technologies and natural language processing (NLP) methods through large language models (LLMs). 
Despite numerous advancements in the NLP research due to Large Language Models (LLMs),  available resources mainly cover ~20 out of the estimated 7,000 languages \cite{magueresse2020lowresource}. As a result, majority of world languages could still be considered as ''low-resource". 

Being a low-resource language (LRL) encompasses different types of inadequacies with respect to the availability of data for creating language technologies \cite{gupta-2022-building}. %In alliance with the theme track of NAACL24, 
Focusing on multilingual linguistic scene in South America, we test the performance of LLMs for annotating part-of-speech (POS) tagging for 12 LRLs from Brazil, make a comparison with 2 LRLs from Africa and 2 high resource languages (HRLs) (e.g., English and Brazilian Portuguese) through human evaluation. 

The evaluation is challenging for two reasons. First, there is a lack of benchmark datasets for the LRLs in Brazil in general. The ones we were able to find in universal dependencies (UD) data base, ~\footnote{\url{https://universaldependencies.org/}} do not have the training data to fine-tune multilingual language models. Hence, we can only leverage prompting LLMs or cross-lingual transfer through multilingual language models. Secondly, there is a lack of large monolingual data to benefit from effective multilingual and cross-lingual transfer techniques~\citep{pfeiffer-etal-2020-mad,ansell-etal-2022-composable,alabi-etal-2022-adapting}. We could only find the Bible corpora with less than 35K sentences for 7 out of the 12 languages. 

We perform the evaluation on 12 Brazilian LRLs by prompting GPT-4 LLM and cross-lingual transfer individually from English and Brazilian Portuguese leveraging XLM-R. We preferred GPT-4 because the other open multilingual models (e.g., mT0~\citep{Muennighoff2022CrosslingualGT}, AYA~\citep{Ustun2024AyaMA}) do not support the LRLs in this study. The results of both methods indicate low performance (less than $34.0\%$  while high-resource languages achieved over $90.0\%$). However, GPT-4 leads to better results and Brazilian Portuguese performs better than English in zero-shot evaluation. Furthermore, to boost the performance of cross-lingual transfer, we perform language adaptation using XLM-R on each language, before fine-tuning Brazilian Portuguese, and evaluating on that language. This boosts the performance by $+3$ to $+12.0$ points on six out of seven languages. Our findings suggest that cross-lingual transfer to these languages is very challenging and having few training examples may further boost the performance. Therefore, there is a need for  building NLP resources across different tasks for these LRLs.
%Two reasons to mention:\\
%Lack of benchmark datasets for LRLs (we have only UD for Brazilian languages)
%Difficult to make progress/improvement in cross-lingual transfer due to lack of data for LRLs.

%The particular type of low-resourced-ness which is of our interest here is “language-specific low-resourced-ness” (Gupta, 2019), which has to do with lack of availability of resources for carrying out NLP tasks, particularly, in our case, automatic annotation of natural language texts.

%In this direction, this paper asks one central question: How good are LLMs resources for low-resource languages, especially on syntactic parsing and part-of-speech (POS) annotations? To provide adequate answers, we have established two specific goals: (i) to evaluate (based on human evaluation) existing LLM and pre-trained models, using 2 Amazonian languages, English, and Brazilian Portuguese languages; (ii) in addition, we will do a comparative study too for pros and cons.
%For this study, we consider four languages: English (Indo-European), Brazilian Portuguese (Indo-European),Karo (Tupi) , and Boróro (Macro-Jê). These four languages represent two extreme points in the continuum of NLP resource availability.
\subsection{Multilingualism in Brazil}
Brazil is the 5th largest country of the world (qua land area) with a population of ~203 million\footnote{Instituto Brasileiro de Geografia e Estatística. 2023. \url{https://www.ibge.gov.br/en/cities-and-states.html}. Accessed: 2023-12-15} %\cite{IBGE2023}
and it is highly multilingual. Although (Brazilian) Portuguese is the official language, there are approx. 160 native/indigenous as well as sign and immigrant languages.\footnote{PIB. 2023. \url{https://pib.socioambiental.org/pt/Linguas}. Accessed: 2023-12-15} %\cite{Povos2023}. 
%The number of native Brazilian languages was estimated to be even twice larger before the Portuguese colonization (starting around 16th century). 

%respect to territorial extension, it is one of the few countries with a continental size. As of 2022, Brazil has a population size of approximately 204 million people . 

%Brazil is far from a homogeneous linguistic area. Rather, it is a highly multilingual country, with a large number of languages in constant contact, not only native languages, but also signed languages, and immigration and inheritance languages. 

%The official language is (Brazilian) Portuguese, making Brazil the only Portuguese-speaking country in South America. Historically, when the Portuguese colonizers first arrived in Brazil, in the 16th century, it is estimated that there were at least twice as many native languages spoken in Brazil than there currently are, whose number is approximately 160\footnote{\url{https://pib.socioambiental.org/pt/Linguas}.}. 

Following \citet{rodrigues1986}, the two macro-language families among Brazilian native languages are Tupi (8 language families, 52 languages), and Macro-Jê (7 language families, 39 languages). There are also several large language families (e.g., Karib (21 languages), Arawak (20 languages), Arawá (7 languages), Tukano, Maku, and Yanomami), six smaller language families to the south of the Amazon river (e.g., Guaikurú (1 language), Nambikwára (3 languages), Txapakura (3 languages), Pano (13 languages), Múra (2 languages), and Matukína (4 languages)) and approx. 10 languages which are not part of any these families. 

These languages share grammatical properties due to family inheritance or areal contact \cite{auikhenvalʹd2002language}. In terms of morphology, most of these languages are polysynthetic, head-marking, and agglutinating with little fusion \cite{auikhenvalʹd1999amazonian, hengeveld2007parts}. In term of syntax, there is quite some variation in terms of word order among these languages\cite{campbell2012typological}.

\section{Literature Overview}
\label{rw}
%\paragraph{NLP resources for Brazilian Language}
 In terms of labelled datasets for Brazilian LRLs, we only found datasets from the UD tasks: ~\citet{fabricio_ferraz_gerardi_2022_6563353} developed for TUDET UD treebanks covering 8 Tupian languages, other languages covered in UD are Apurina~\citep{Hamalainen2021-tr}, Bororo, Madi-Jarawara, and Xavante (contributed by the TUDET team). For the monolingual data, we found seven Bible corpora on the eBible corpus~\citep{Akerman2023TheEC} that are freely available. All languages lack a large monolingual corpus which makes it very challenging for cross-lingual transfer and multilingual pre-training of LLMs. 
%There is a growing effort to deal with low-resource languages in the domain of NLP, and, more recently, LLMs have entered the scene as possible suitors to enhance the resources for these languages.

%\paragraph{Evaluation of LLM on non-English languages} ChatGPT Beyond English: Towards a Comprehensive Evaluation of Large Language Models in Multilingual Learning \cite{lai2023chatgpt}, they cover UPOS
In terms of evaluation, some studies have already shown the potential of prompting LLMs in multilingual settings~\citep{ahuja-etal-2023-mega,lai-etal-2023-chatgpt}, including some LRLs~\citep{ojo2023good,ahuja2023megaverse}. However, evaluations covering Brazilian LRLs are lacking. To the best of our knowledge, our study is the first to fill this gap. 

\begin{table*}[t]
\begin{center}
%\resizebox{\textwidth}{!}{%
\scalebox{0.65}{
\begin{tabular}{llr|crrrr}
\toprule
\textbf{Language} &\textbf{Language family} &\textbf{Monolingual data size} & \textbf{UD dataset name} & \textbf{Train} &\textbf{Dev} & \textbf{Test set A}&\textbf{Test set B}  \\
\midrule
\multicolumn{4}{l}{\texttt{high-resource languages}} \\
English (en) & Indo-European/West Germanic & not collected & en\_ewt & 12,544 & 2,001 & 2,007 & -\\
Portuguese (pt) & Indo-European/Romance &  not collected & pt\_gsd & 9,616 & 1,204 & 1,200 &  -\\
\midrule
\multicolumn{4}{l}{\texttt{Brazilian languages}} \\
Apurina (apu)  & Arawakan & Bible (8,729)& apu\_ufpa & - & - & 152 & 134 \\
Akuntsu (aqz) & Tupian &  N/A  & aqz\_tudet & - & - & 343  & 267 \\
Karo (arr) & 	Tupian &  N/A & arr\_tudet & - & - & 674 & 172 \\
Bororo (bor) &	
Macro-Jê &  Bible (8,254) & bor\_bdt & - & - & 371 & 161 \\
Guajajara (gub) & Tupian &  Bible (33,757) & gub\_tudet & - & - & 1,182 & 914\\
Madi-Jarawara (jaa) & 	
Arawan &  Bible (8,606) & jaa\_jarawara & - & - & 20 & 18\\
Makurap (mpu) & Tupian &  N/A & mpu\_tudet & - & - & 37 & 8 \\
Munduruku(myu) &  	
Tupian &  Bible (8,430) & myu\_tudet & - & - & 158 & 82\\
Tupinamba (tpn) & 	
Tupian &  N/A & tpn\_tudet & - & - & 581 & 458 \\
Kaapor (urb) & 	
Tupian &  Bible (8,535)  & urb\_tudet & - & - & 83 & 20\\
Xavante (xav) & 	
Macro-Jê &  Bible (8,213) & xav\_xdt & - & - & 148 & 128 \\
Nheengatu (yrl) & 	
Tupian &  N/A & yrl\_complin & - & - & 1239 & -\\
\midrule
\multicolumn{4}{l}{\texttt{African languages}} \\
Wolof (wo) & Niger-Congo/Senegambian &  not collected & wo\_wtb & 1188 & 449 & 470 & 470 \\
Yoruba (yo) & Niger-Congo/Volta-Niger &  not collected &  yo\_ytb & - & - & 318 & 318\\
\bottomrule
\end{tabular}
}
\caption{\textbf{UD-POS datasets in our evaluation:} We provide the training, validation and test splits we used for experiments. Test set A are the original test set in UD, the Test set B is a subset of A where we removed sentences that GPT-4 is not able to run inference for due to non-identification of the language. }
\label{ud_pos_data}
\vspace{-4mm}
\end{center}
\end{table*}

\begin{table*}[t]
\addtolength{\tabcolsep}{-1pt}
\scriptsize
\begin{center}
\resizebox{\textwidth}{!}{%
\begin{tabular}{lr|rrrrrr|r}
\toprule
 &\textbf{XLM-R} & \multicolumn{6}{c}{\textbf{XLM-R (zero-shot cross-lingual transfer)}} & \textbf{GPT-4} \\
   \textbf{Language}&\textbf{Test set A} &\multicolumn{3}{c}{\textbf{Test set A}} &\multicolumn{3}{c}{\textbf{Test set B}} & \textbf{Test set B} \\

 &\textbf{Full-sup.} &\textbf{en$\rightarrow$ xx} & \textbf{pt$\rightarrow$ xx} & \textbf{LAFT + pt$\rightarrow$ xx} &\textbf{en$\rightarrow$ xx} & \textbf{pt$\rightarrow$ xx} & \textbf{LAFT + pt$\rightarrow$ xx}& \textbf{0-shot} \\
\midrule
\multicolumn{2}{l}{\texttt{high-resource languages}} \\
en\_ewt & \textbf{98.0} & \textbf{98.0} & 83.6 & &  & 91.9\\
pt\_gsd & \textbf{97.8} & 90.0 & \textbf{97.8 }& & & 92.4\\
\midrule
\multicolumn{2}{l}{\texttt{Brazilian languages}} \\
apu\_ufpa & - & 37.5 & 40.6 & \textbf{44.9} & 36.8 & 40.2 & \textbf{44.7} & 42.6\\
aqz\_tudet & - & 31.9 & 37.8 & & 31.3 & 36.8 & & \textbf{49.5} \\
arr\_tudet & - & 3.9 & 14.9  & & 6.3 & 19.8 & & \textbf{27.7} \\
bor\_bdt & - & 19.0 & 23.5 & 27.3 & 18.4 & 23.0 & 26.4 & \textbf{41.3} \\
gub\_tudet & - & 26.5 & 30.2  & 36.0 & 27.8 & 32.1 & \textbf{37.1} & 36.2 \\
jaa\_jarawara & - & 28.2 & 28.4 &\textbf{34.5} & 27.2 & 27.9 & 33.6 & \textbf{33.0}\\
mpu\_tudet & - & 4.9& \textbf{9.0} && 0.0 & 0.8 & & 0.0 \\
myu\_tudet & - & 21.2 &27.1& \textbf{30.3} & 10.8 & 14.8 & 16.5 & 18.2 \\
tpn\_tudet & - & 39.1 & 41.9 && 38.9 & 41.8 & & \textbf{47.2}\\
urb\_tudet & - & 7.8 & 11.8 &21.2 & 9.2 &  9.5& 21.6 & \textbf{32.3} \\
xav\_xdt & - & 26.5 & 29.0 & 28.2& 27.3 & 29.9 & 29.3 & \textbf{36.5} \\
yrl\_complin & - & 28.9 & 31.5 &  & 29.0 & 31.7 &  & \textbf{41.2} \\
\midrule
\multicolumn{2}{l}{\texttt{African languages}} \\
wo\_wtb &  \textbf{87.6} & 29.3 & 35.6& & & && 64.8\\
yo\_ytb &  - & 22.5 & 31.5 & & &  && \textbf{75.4} \\
\midrule
Average (Brazilian languages) &  - & 23.0 & 27.1 & & 21.9 & 25.7 & &  33.8 \\

\bottomrule
\end{tabular}
}
\caption{\textbf{POS accuracy results for Brazilian languages:} We compare the accuracy of GPT-4 to zero-shot cross-lingual transfer from English language and Portuguese leveraging XLM-R-large multilingual pre-trained language model. Test set A is the original test set found on UD while Test set B are the ones GPT-4 could automatically detect their language to run inference. }
\vspace{-3mm}
\label{pos_result}
\end{center}
\end{table*}

\section{Experimental setup}
\label{exs}
We focus our evaluation of POS tagging (a subtask of universal dependencies (UD)) on Brazilian LRLs due to the simplicity of the task, its popularity, and  the availability of the test evaluation datasets in UD .\footnote{\url{https://universaldependencies.org/}} %We describe the evaluation dataset and the models used in the subsections below.  
\subsection{Evaluation Datasets}
\label{data}
We evaluated 12 Brazilian LRLs and 2 African languages for a comparison across other regions with low-resource languages. Finally, we added 2 HRLs (i.e., English and Brazilian Portuguese). Our definition of HRL is based on the size of unlabelled data on the web. The larger their size are, the more likely they are included in pre-training of the LLMs~\footnote{\url{https://help.openai.com/en/articles/8357869-chatgpt-language-support-alpha-web}} and multilingual pre-trained LMs~\citep{conneau-etal-2020-unsupervised}.    %~\cite{11234/1-5287}
While  UD~\cite{11234/1-5287} covers many languages, most LRLs only have a test set because of their limited sizes (less than 10k tokens). The Brazilian LRLs we evaluated on have also less than 13k tokens (except Nheengatu with 12,621 tokens).  

\autoref{ud_pos_data} shows the languages in our evaluation, their language family, availability of monolingual corpus or Bible corpus in that language, UD dataset, and sizes. We collected the Bible corpus from the eBible website and used it for language adaptation. We have two test sets in our evaluation: (1) \textbf{Test set A}: the original test set in the UD benchmark (2) \textbf{Test set B} the subsample of Test set A where we removed sentences that GPT-4 fails to provide predictions for (mostly due to not properly identifying the language). We added this information for a fair comparison of the methods (i.e. using the same number of sentences in evaluation).

\subsection{Models}
For the experiments, we consider three approaches that are popular in the zero-shot setting since we lack the training data for the Brazilian languages (see \autoref{appendix_model} for details). 

\paragraph{Prompting GPT-4} %GPT-4~\footnote{\url{https://chat.openai.com/}} is a large language model developed by pre-training on a large amount of texts and code from the web, followed by instruction prompt tuning based on human feedback.
We prompt GPT-4 using a similar prompt provided by \citet{lai-etal-2023-chatgpt} where the model is provided a task description before the input (see\autoref{appendix_prompt} for details).

\paragraph{Cross-lingual transfer} We trained a POS tagger individually for English and Portuguese, and perform the zero-shot transfer on other languages. We used the XLM-R-large (or simply, XLM-R)~\citep{conneau-etal-2020-unsupervised} for training the models. %XLM-R has been pre-trained on 100 languages of the world with over 2TB pre-training corpus size but this corpus does not include any indigenous Brazilian languages. 

\paragraph{Language Adaptive Fine-tuning (LAFT)} We leverage LAFT for an effective cross-lingual transfer by first adapting XLM-R-large model to a new language with limited amount of monolingual data~\citep{alabi-etal-2020-massive,pfeiffer-etal-2020-mad,chau-smith-2021-specializing,alabi-etal-2022-adapting}. %This method was proven to be very effective for low-resource languages~\citep{adelani-etal-2021-masakhaner,muller-etal-2021-unseen}. 
We make use of the Bible data as the fine-tuning corpus since it is the largest one for these languages and we only found 7 (out of 12 Brazilian languages) languages which have a Bible corpus. Similar to \citet{ebrahimi-kann-2021-adapt}, we examine the effectiveness of this small pre-training corpus with 8K-34K sentences. According to \citet{pfeiffer-etal-2020-mad}, this approach can significantly boost cross-lingual transfer. However, it is not parameter-efficient like the MAD-X they proposed. On the other hand, \citet{ebrahimi-kann-2021-adapt} argued that simple adaptation to a new language is more effective than MAD-X especially when using the Bible corpus for adaptation and we follow this recommendation in our evaluation. 

%\paragraph{Hyper-parameter of experiments} For the cross-lingual and LAFT experiments, we used HuggingFace transformers~\citep{wolf-etal-2020-transformers} and A100 Nvidia GPU for fine-tuning the models.  %while for GPT-4 we use the Open AI API. For the LAFT, we train for 3 epochs on one GPU while for cross-lingual, we fine-tune English and Portuguese individually using a batch size of 64, with gradient accumulation of 2, and a training epoch of 10. 

\section{Results}
\autoref{pos_result} shows the result of our evaluation on POS tagging with the following key findings: 
\paragraph{Zero-shot evaluation results} While POS tagging has a performance of 98\% (e.g. for English and Portuguese) when training data are available (especially for HRLs), the performance decreases while performing zero-shot transfer to other languages because POS tagging is language-specific. The transfer performance %from English and Portuguese 
is low for both Brazilian and African languages (probably) because they are not typologically related whereas English and Portuguese are slightly related (i.e., being in the same Indo-European family) and covered by XLM-R, thus achieving an impressive transfer performance ($>+83\%$). 

\paragraph{GPT-4 vs. basic cross-lingual transfer}  GPT-4 performed slightly better than the zero-shot transfer from other languages in our experiments indicating better abilities of LLMs for this task. For English and Portuguese, the performance reaches to $90\%$ (although it is not on par with fully-supervised setting). For African languages, the performance was lower than the HRLs, but it was still decent ($64.8$-$75.4$) probably because the LLMs were exposed to some African languages during pre-training. The struggle of GPT-4 for Brazilian LRLs can be explained with the fact that these languages were probably not included during the pre-training. The generation is often not useful for some examples, where GPT-4 declines to give answers like ``\textit{As an AI, I'm unable to provide the POS tags for words in languages I'm not programmed to understand. }''. Thus, we had to remove such examples from our evaluation. However, this was not the case for African LRLs and the HRLs. 

\paragraph{Language adaptation for cross-lingual transfer performance} We performed LAFT training %for 3 epochs 
on the Bible corpus individually for the \textit{apu}, \textit{bor}, \textit{gub}, \textit{jaa}, \textit{myu}, \textit{urb}, and \textit{xav}. Our results indicate an improvement in accuracy on 6 out of the 7 languages, except for \textit{xav}. The performance improvement is quite large for \textit{urb} (+7.2 on test A, and 12.1 on Test B), and moderate improvement of $+3$ to $+6$ for other languages. This experimental result shows that with sufficient monolingual texts, we can increase the performance of the cross-lingual transfer results. However, for the LRLs, such data is scarce. A more effective approach is perhaps to annotate few examples (e.g. 10 or 100 sentences) for training POS taggers %for these languages 
to boost the performance (cf. ~\citep{lauscher-etal-2020-zero,hedderich-etal-2020-transfer} for a larger boost in performance for token classification tasks in this few-short setting). Regardless, there is a need for better methods to leverage small monolingual data sets.  %Alternatively, better methods that can leverage small monolingual data needs to be developed. %for these languages. 
%In the next section, we will elaborate on the on the error analysis.  

\section{Error analysis}

%One type of error in the GPT analysis could possibly be explained with reference to the form of the word. Consider the following examples from Karo and Guajajara, respectively. 

In this section, we provide examples from 2 Brazilian languages (Karo and Guajajara) where the LLMs made errors with the POS tagging. The first line refers to the original sentence, the second line refers to the gold-standard UD POS tag; the third line refers to the GPT-4 POS tag.)

In example (1), the auxiliary verb (in Karo) has the same orthographic form as the English interjection \textit{okay}. In example (2), the Guajajara verb has (partially) the same orthographic form as the English interjection (\textit{oh}). Due to these similarities, GPT-4 seems to tag the POS for these words according to English instead of the POS tagged in UD for Karo and Guajajara.

\begin{comment}
\begin{enumerate}
\item  aʔwero&  toba& \textbf{okay}\\ 
NOUN& VERB& \textbf{AUX}\\
NOUN&  NOUN& \textbf{INTJ}\\
\vspace{-7mm}
\item \textbf{Oho}&  kaʔapiʔi&  rehe&. \\ 
\textbf{VERB}&  NOUN&  ADP&PUNCT\\
\textbf{INT}& VERB& ADV&PUNCT\\  
\end{enumerate}
\end{comment}

\begin{enumerate}
\item  awero   toba  \textbf{okay}\\ 
NOUN VERB \textbf{AUX} \\
NOUN   NOUN  \textbf{INTJ}\\
\vspace{-7mm}
\item \textbf{Oho}  kaapii  rehe . \\ 
\textbf{VERB}  NOUN  ADP PUNCT\\
\textbf{INT} \ \ \ \ VERB \ ADV PUNCT\\  
\end{enumerate}

\section{Discussion \& Conclusions}
In our study, we explored how LLMs perform the NLP task of POS tagging for 12 LRLs in Brazil and compared this performance with 2 LRLs in Africa and 2 HRLs (English, Brazilian Portuguese). POS is a well established NLP task and it provides insights about the linguistic structures of the different languages especially when only limited data is available, such linguistic annotations have been shown to improve language understanding and generation for endangered languages~\citep{Zhang2024HireAL}. Our results indicate that the LLMs (GPT-4) perform worse for LRLs on this task in general but older approaches like language adaptive fine-tuning that leverage multilingual encoder models provides some improvements.  However, with the lack of available data, any improvements across methods are limited. 
%also not feasible. 
Although we focused on 12 Brazilian LRLs, there are many other LRLs which we were not able to cover. Future work can expand this evaluation to more tasks and to other LRLs not only from Brazil but from other regions around the world as well.

\section{Limitations}
%As an ongoing work to evaluate ChatGPT and LLMs on multilingual learning tasks, our current work observes several limitations that can be addressed in future studies. First, although our experiments have covered 37 languages, including lowand extremely low-languages, there are still many other languages that are not explored in the current work. Some tasks/datasets in our work have not covered lower-resource languages. 

Due to limited space, we only focused on POS tagging for this paper but there is a need to explore how LLMs perform other NLP tasks for LRLs. We only evaluated ChatGPT in the zero-shot learning setting but we do not have comparisons with other recent multilingual LLMs, e.g., BLOOM (Scao et al., 2022), and Gemini, in various other learning scenarios. While some of these models are currently less accessible for large-scale evaluations, our plan is to include more models and learning settings along the way to strengthen our evaluations and comparisons in the future. Finally, the current work only evaluates ChatGPT in terms of performance over NLP tasks in different languages. To better characterize ChatGPT and LLMs, other evaluation metrics should also be investigated to report more complete perspectives for multilingual learning, including but not limited to adversarial robustness, biases, toxic/harmful content, hallucination, accessibility, development costs, and interoperability.

%including those with available multilingual datasets, have not been considered in the current work. Examining more tasks and datasets will enable a more comprehensive understanding of ChatGPT and LLMs in multilingual settings. Third, our current work only evaluates 

\section {Ethics Issues}
Since we used publicly available data sets, we do not foresee any major issues in terms of ethical concerns. 
%\section{Conclusions}
\section*{Acknowledgements}

Atul Kr. Ojha would like to acknowledge the support of the Science Foundation Ireland (SFI) as part of Grant Number SFI/12/RC/2289\_P2 Insight\_2, Insight SFI Centre for Data Analytics and CA21167 COST Action UniDive (by COST (European Cooperation in Science and Technology). David Adelani acknowledges the support of DeepMind Academic Fellowship programme.
\bibliography{anthology,naacl}

\begin{thebibliography}{30}
\expandafter\ifx\csname natexlab\endcsname\relax\def\natexlab#1{#1}\fi

\bibitem[{Adelani et~al.(2021)Adelani, Abbott, Neubig, D{'}souza, Kreutzer,
  Lignos, Palen-Michel, Buzaaba, Rijhwani, Ruder, Mayhew, Azime, Muhammad,
  Emezue, Nakatumba-Nabende, Ogayo, Anuoluwapo, Gitau, Mbaye, Alabi, Yimam,
  Gwadabe, Ezeani, Niyongabo, Mukiibi, Otiende, Orife, David, Ngom, Adewumi,
  Rayson, Adeyemi, Muriuki, Anebi, Chukwuneke, Odu, Wairagala, Oyerinde, Siro,
  Bateesa, Oloyede, Wambui, Akinode, Nabagereka, Katusiime, Awokoya, MBOUP,
  Gebreyohannes, Tilaye, Nwaike, Wolde, Faye, Sibanda, Ahia, Dossou, Ogueji,
  DIOP, Diallo, Akinfaderin, Marengereke, and
  Osei}]{adelani-etal-2021-masakhaner}
David~Ifeoluwa Adelani, Jade Abbott, Graham Neubig, Daniel D{'}souza, Julia
  Kreutzer, Constantine Lignos, Chester Palen-Michel, Happy Buzaaba, Shruti
  Rijhwani, Sebastian Ruder, Stephen Mayhew, Israel~Abebe Azime, Shamsuddeen~H.
  Muhammad, Chris~Chinenye Emezue, Joyce Nakatumba-Nabende, Perez Ogayo, Aremu
  Anuoluwapo, Catherine Gitau, Derguene Mbaye, Jesujoba Alabi, Seid~Muhie
  Yimam, Tajuddeen~Rabiu Gwadabe, Ignatius Ezeani, Rubungo~Andre Niyongabo,
  Jonathan Mukiibi, Verrah Otiende, Iroro Orife, Davis David, Samba Ngom, Tosin
  Adewumi, Paul Rayson, Mofetoluwa Adeyemi, Gerald Muriuki, Emmanuel Anebi,
  Chiamaka Chukwuneke, Nkiruka Odu, Eric~Peter Wairagala, Samuel Oyerinde,
  Clemencia Siro, Tobius~Saul Bateesa, Temilola Oloyede, Yvonne Wambui, Victor
  Akinode, Deborah Nabagereka, Maurice Katusiime, Ayodele Awokoya, Mouhamadane
  MBOUP, Dibora Gebreyohannes, Henok Tilaye, Kelechi Nwaike, Degaga Wolde,
  Abdoulaye Faye, Blessing Sibanda, Orevaoghene Ahia, Bonaventure F.~P. Dossou,
  Kelechi Ogueji, Thierno~Ibrahima DIOP, Abdoulaye Diallo, Adewale Akinfaderin,
  Tendai Marengereke, and Salomey Osei. 2021.
\newblock \href {https://doi.org/10.1162/tacl_a_00416} {{M}asakha{NER}: Named
  entity recognition for {A}frican languages}.
\newblock \emph{Transactions of the Association for Computational Linguistics},
  9:1116--1131.

\bibitem[{Ahuja et~al.(2023{\natexlab{a}})Ahuja, Diddee, Hada, Ochieng, Ramesh,
  Jain, Nambi, Ganu, Segal, Ahmed, Bali, and Sitaram}]{ahuja-etal-2023-mega}
Kabir Ahuja, Harshita Diddee, Rishav Hada, Millicent Ochieng, Krithika Ramesh,
  Prachi Jain, Akshay Nambi, Tanuja Ganu, Sameer Segal, Mohamed Ahmed, Kalika
  Bali, and Sunayana Sitaram. 2023{\natexlab{a}}.
\newblock \href {https://aclanthology.org/2023.emnlp-main.258} {{MEGA}:
  Multilingual evaluation of generative {AI}}.
\newblock In \emph{Proceedings of the 2023 Conference on Empirical Methods in
  Natural Language Processing}, pages 4232--4267, Singapore. Association for
  Computational Linguistics.

\bibitem[{Ahuja et~al.(2023{\natexlab{b}})Ahuja, Aggarwal, Gumma, Watts, Sathe,
  Ochieng, Hada, Jain, Axmed, Bali et~al.}]{ahuja2023megaverse}
Sanchit Ahuja, Divyanshu Aggarwal, Varun Gumma, Ishaan Watts, Ashutosh Sathe,
  Millicent Ochieng, Rishav Hada, Prachi Jain, Maxamed Axmed, Kalika Bali,
  et~al. 2023{\natexlab{b}}.
\newblock Megaverse: Benchmarking large language models across languages,
  modalities, models and tasks.
\newblock \emph{arXiv preprint arXiv:2311.07463}.

\bibitem[{Aikhenvald(2002)}]{auikhenvalʹd2002language}
Alexandra~Y. Aikhenvald. 2002.
\newblock \emph{Language contact in Amazonia}.
\newblock Oxford University Press.

\bibitem[{Akerman et~al.(2023)Akerman, Baines, Daspit, Hermjakob, Jang, Leong,
  Martin, Mathew, Robie, and Schwarting}]{Akerman2023TheEC}
Vesa Akerman, David Baines, Damien Daspit, Ulf Hermjakob, Tae~Young Jang, Colin
  Leong, Michael Martin, Joel Mathew, Jonathan Robie, and Marcus Schwarting.
  2023.
\newblock \href {https://api.semanticscholar.org/CorpusID:258236091} {The
  ebible corpus: Data and model benchmarks for bible translation for
  low-resource languages}.
\newblock \emph{ArXiv}, abs/2304.09919.

\bibitem[{Alabi et~al.(2020)Alabi, Amponsah-Kaakyire, Adelani, and
  Espa{\~n}a-Bonet}]{alabi-etal-2020-massive}
Jesujoba Alabi, Kwabena Amponsah-Kaakyire, David Adelani, and Cristina
  Espa{\~n}a-Bonet. 2020.
\newblock \href {https://aclanthology.org/2020.lrec-1.335} {Massive vs. curated
  embeddings for low-resourced languages: the case of {Y}or{\`u}b{\'a} and
  {T}wi}.
\newblock In \emph{Proceedings of the Twelfth Language Resources and Evaluation
  Conference}, pages 2754--2762, Marseille, France. European Language Resources
  Association.

\bibitem[{Alabi et~al.(2022)Alabi, Adelani, Mosbach, and
  Klakow}]{alabi-etal-2022-adapting}
Jesujoba~O. Alabi, David~Ifeoluwa Adelani, Marius Mosbach, and Dietrich Klakow.
  2022.
\newblock \href {https://aclanthology.org/2022.coling-1.382} {Adapting
  pre-trained language models to {A}frican languages via multilingual adaptive
  fine-tuning}.
\newblock In \emph{Proceedings of the 29th International Conference on
  Computational Linguistics}, pages 4336--4349, Gyeongju, Republic of Korea.
  International Committee on Computational Linguistics.

\bibitem[{Ansell et~al.(2022)Ansell, Ponti, Korhonen, and
  Vuli{\'c}}]{ansell-etal-2022-composable}
Alan Ansell, Edoardo Ponti, Anna Korhonen, and Ivan Vuli{\'c}. 2022.
\newblock \href {https://doi.org/10.18653/v1/2022.acl-long.125} {Composable
  sparse fine-tuning for cross-lingual transfer}.
\newblock In \emph{Proceedings of the 60th Annual Meeting of the Association
  for Computational Linguistics (Volume 1: Long Papers)}, pages 1778--1796,
  Dublin, Ireland. Association for Computational Linguistics.

\bibitem[{Campbell(2012)}]{campbell2012typological}
Lyle Campbell. 2012.
\newblock Typological characteristics of south american indigenous languages.
\newblock \emph{The indigenous languages of South America: A comprehensive
  guide}, pages 259--330.

\bibitem[{Chau and Smith(2021)}]{chau-smith-2021-specializing}
Ethan~C. Chau and Noah~A. Smith. 2021.
\newblock \href {https://doi.org/10.18653/v1/2021.mrl-1.5} {Specializing
  multilingual language models: An empirical study}.
\newblock In \emph{Proceedings of the 1st Workshop on Multilingual
  Representation Learning}, pages 51--61, Punta Cana, Dominican Republic.
  Association for Computational Linguistics.

\bibitem[{Conneau et~al.(2020)Conneau, Khandelwal, Goyal, Chaudhary, Wenzek,
  Guzm{\'a}n, Grave, Ott, Zettlemoyer, and
  Stoyanov}]{conneau-etal-2020-unsupervised}
Alexis Conneau, Kartikay Khandelwal, Naman Goyal, Vishrav Chaudhary, Guillaume
  Wenzek, Francisco Guzm{\'a}n, Edouard Grave, Myle Ott, Luke Zettlemoyer, and
  Veselin Stoyanov. 2020.
\newblock \href {https://doi.org/10.18653/v1/2020.acl-main.747} {Unsupervised
  cross-lingual representation learning at scale}.
\newblock In \emph{Proceedings of the 58th Annual Meeting of the Association
  for Computational Linguistics}, pages 8440--8451, Online. Association for
  Computational Linguistics.

\bibitem[{Dixon and Aikhenvald(1999)}]{auikhenvalʹd1999amazonian}
R.~M.~W. Dixon and Alexandra~Y. Aikhenvald. 1999.
\newblock \emph{The Amazonian Languages}.
\newblock Cambridge University Press.

\bibitem[{Ebrahimi and Kann(2021)}]{ebrahimi-kann-2021-adapt}
Abteen Ebrahimi and Katharina Kann. 2021.
\newblock \href {https://doi.org/10.18653/v1/2021.acl-long.351} {How to adapt
  your pretrained multilingual model to 1600 languages}.
\newblock In \emph{Proceedings of the 59th Annual Meeting of the Association
  for Computational Linguistics and the 11th International Joint Conference on
  Natural Language Processing (Volume 1: Long Papers)}, pages 4555--4567,
  Online. Association for Computational Linguistics.

\bibitem[{Gerardi et~al.(2022)Gerardi, Reichert, Aragon, Martín-Rodríguez,
  Godoy, and Merzhevich}]{fabricio_ferraz_gerardi_2022_6563353}
Fabrício~Ferraz Gerardi, Stanislav Reichert, Carolina Aragon, Lorena
  Martín-Rodríguez, Gustavo Godoy, and Tatiana Merzhevich. 2022.
\newblock \href {https://doi.org/10.5281/zenodo.6563353} {Tudet: Tupían
  dependency treebank}.

\bibitem[{Gupta(2022)}]{gupta-2022-building}
Akshat Gupta. 2022.
\newblock \href {https://doi.org/10.18653/v1/2022.sigmorphon-1.1} {On building
  spoken language understanding systems for low resourced languages}.
\newblock In \emph{Proceedings of the 19th SIGMORPHON Workshop on Computational
  Research in Phonetics, Phonology, and Morphology}, pages 1--11, Seattle,
  Washington. Association for Computational Linguistics.

\bibitem[{H{\"a}m{\"a}l{\"a}inen et~al.(2021)H{\"a}m{\"a}l{\"a}inen,
  {University of Helsinki}, Partanen, and Alnajjar}]{Hamalainen2021-tr}
Mika H{\"a}m{\"a}l{\"a}inen, {University of Helsinki}, Niko Partanen, and
  Khalid Alnajjar, editors. 2021.
\newblock \emph{Multilingual Facilitation}.
\newblock University of Helsinki.

\bibitem[{Hedderich et~al.(2020)Hedderich, Adelani, Zhu, Alabi, Markus, and
  Klakow}]{hedderich-etal-2020-transfer}
Michael~A. Hedderich, David Adelani, Dawei Zhu, Jesujoba Alabi, Udia Markus,
  and Dietrich Klakow. 2020.
\newblock \href {https://doi.org/10.18653/v1/2020.emnlp-main.204} {Transfer
  learning and distant supervision for multilingual transformer models: A study
  on {A}frican languages}.
\newblock In \emph{Proceedings of the 2020 Conference on Empirical Methods in
  Natural Language Processing (EMNLP)}, pages 2580--2591, Online. Association
  for Computational Linguistics.

\bibitem[{Hengeveld et~al.(2007)}]{hengeveld2007parts}
Kees Hengeveld et~al. 2007.
\newblock Parts-of-speech systems and morphological types.
\newblock \emph{ACLC Working Papers Volume 2, issue}, page~31.

\bibitem[{Lai et~al.(2023)Lai, Ngo, Pouran Ben~Veyseh, Man, Dernoncourt, Bui,
  and Nguyen}]{lai-etal-2023-chatgpt}
Viet Lai, Nghia Ngo, Amir Pouran Ben~Veyseh, Hieu Man, Franck Dernoncourt,
  Trung Bui, and Thien Nguyen. 2023.
\newblock \href {https://aclanthology.org/2023.findings-emnlp.878} {{C}hat{GPT}
  beyond {E}nglish: Towards a comprehensive evaluation of large language models
  in multilingual learning}.
\newblock In \emph{Findings of the Association for Computational Linguistics:
  EMNLP 2023}, pages 13171--13189, Singapore. Association for Computational
  Linguistics.

\bibitem[{Lauscher et~al.(2020)Lauscher, Ravishankar, Vuli{\'c}, and
  Glava{\v{s}}}]{lauscher-etal-2020-zero}
Anne Lauscher, Vinit Ravishankar, Ivan Vuli{\'c}, and Goran Glava{\v{s}}. 2020.
\newblock \href {https://doi.org/10.18653/v1/2020.emnlp-main.363} {From zero to
  hero: {O}n the limitations of zero-shot language transfer with multilingual
  {T}ransformers}.
\newblock In \emph{Proceedings of the 2020 Conference on Empirical Methods in
  Natural Language Processing (EMNLP)}, pages 4483--4499, Online. Association
  for Computational Linguistics.

\bibitem[{Magueresse et~al.(2020)Magueresse, Carles, and
  Heetderks}]{magueresse2020lowresource}
Alexandre Magueresse, Vincent Carles, and Evan Heetderks. 2020.
\newblock \href {http://arxiv.org/abs/2006.07264} {Low-resource languages: A
  review of past work and future challenges}.

\bibitem[{Muennighoff et~al.(2022)Muennighoff, Wang, Sutawika, Roberts,
  Biderman, Scao, Bari, Shen, Yong, Schoelkopf, Tang, Radev, Aji, Almubarak,
  Albanie, Alyafeai, Webson, Raff, and Raffel}]{Muennighoff2022CrosslingualGT}
Niklas Muennighoff, Thomas Wang, Lintang Sutawika, Adam Roberts, Stella
  Biderman, Teven~Le Scao, M~Saiful Bari, Sheng Shen, Zheng~Xin Yong, Hailey
  Schoelkopf, Xiangru Tang, Dragomir~R. Radev, Alham~Fikri Aji, Khalid
  Almubarak, Samuel Albanie, Zaid Alyafeai, Albert Webson, Edward Raff, and
  Colin Raffel. 2022.
\newblock \href {https://api.semanticscholar.org/CorpusID:260641062}
  {Crosslingual generalization through multitask finetuning}.
\newblock \emph{ArXiv}, abs/2211.01786.

\bibitem[{Muller et~al.(2021)Muller, Anastasopoulos, Sagot, and
  Seddah}]{muller-etal-2021-unseen}
Benjamin Muller, Antonios Anastasopoulos, Beno{\^\i}t Sagot, and Djam{\'e}
  Seddah. 2021.
\newblock \href {https://doi.org/10.18653/v1/2021.naacl-main.38} {When being
  unseen from m{BERT} is just the beginning: Handling new languages with
  multilingual language models}.
\newblock In \emph{Proceedings of the 2021 Conference of the North American
  Chapter of the Association for Computational Linguistics: Human Language
  Technologies}, pages 448--462, Online. Association for Computational
  Linguistics.

\bibitem[{Ojo et~al.(2023)Ojo, Ogueji, Stenetorp, and Adelani}]{ojo2023good}
Jessica Ojo, Kelechi Ogueji, Pontus Stenetorp, and David~I Adelani. 2023.
\newblock How good are large language models on african languages?
\newblock \emph{arXiv preprint arXiv:2311.07978}.

\bibitem[{Pfeiffer et~al.(2020)Pfeiffer, Vuli{\'c}, Gurevych, and
  Ruder}]{pfeiffer-etal-2020-mad}
Jonas Pfeiffer, Ivan Vuli{\'c}, Iryna Gurevych, and Sebastian Ruder. 2020.
\newblock \href {https://doi.org/10.18653/v1/2020.emnlp-main.617} {{MAD-X}:
  {A}n {A}dapter-{B}ased {F}ramework for {M}ulti-{T}ask {C}ross-{L}ingual
  {T}ransfer}.
\newblock In \emph{Proceedings of the 2020 Conference on Empirical Methods in
  Natural Language Processing (EMNLP)}, pages 7654--7673, Online. Association
  for Computational Linguistics.

\bibitem[{Rodrigues(1986)}]{rodrigues1986}
Aryon~Dall'lgna Rodrigues. 1986.
\newblock \emph{Línguas Brasileiras: para o conhecimento das línguas
  indígenas}.
\newblock São Paulo: Edições Loyola.

\bibitem[{Ustun et~al.(2024)Ustun, Aryabumi, Yong, Ko, D'souza, Onilude,
  Bhandari, Singh, Ooi, Kayid, Vargus, Blunsom, Longpre, Muennighoff, Fadaee,
  Kreutzer, and Hooker}]{Ustun2024AyaMA}
A.~Ustun, Viraat Aryabumi, Zheng-Xin Yong, Wei-Yin Ko, Daniel D'souza,
  Gbemileke Onilude, Neel Bhandari, Shivalika Singh, Hui-Lee Ooi, Amr Kayid,
  Freddie Vargus, Phil Blunsom, Shayne Longpre, Niklas Muennighoff, Marzieh
  Fadaee, Julia Kreutzer, and Sara Hooker. 2024.
\newblock \href {https://api.semanticscholar.org/CorpusID:267627803} {Aya
  model: An instruction finetuned open-access multilingual language model}.
\newblock \emph{ArXiv}, abs/2402.07827.

\bibitem[{Wolf et~al.(2020)Wolf, Debut, Sanh, Chaumond, Delangue, Moi, Cistac,
  Rault, Louf, Funtowicz, Davison, Shleifer, von Platen, Ma, Jernite, Plu, Xu,
  Le~Scao, Gugger, Drame, Lhoest, and Rush}]{wolf-etal-2020-transformers}
Thomas Wolf, Lysandre Debut, Victor Sanh, Julien Chaumond, Clement Delangue,
  Anthony Moi, Pierric Cistac, Tim Rault, Remi Louf, Morgan Funtowicz, Joe
  Davison, Sam Shleifer, Patrick von Platen, Clara Ma, Yacine Jernite, Julien
  Plu, Canwen Xu, Teven Le~Scao, Sylvain Gugger, Mariama Drame, Quentin Lhoest,
  and Alexander Rush. 2020.
\newblock \href {https://doi.org/10.18653/v1/2020.emnlp-demos.6} {Transformers:
  State-of-the-art natural language processing}.
\newblock In \emph{Proceedings of the 2020 Conference on Empirical Methods in
  Natural Language Processing: System Demonstrations}, pages 38--45, Online.
  Association for Computational Linguistics.

\bibitem[{Zeman et~al.(2023)Zeman, Nivre, Abrams, Ackermann, Aepli, Aghaei,
  Agi{\'c}, Ahmadi, Ahrenberg, Ajede, Akkurt, Aleksandravi{\v c}i{\=u}t{\.e},
  Alfina, Algom, Alnajjar, Alzetta, Andersen, Antonsen, Aoyama, Aplonova,
  Aquino, Aragon, Aranes, Aranzabe, and et~al}]{11234/1-5287}
Daniel Zeman, Joakim Nivre, Mitchell Abrams, Elia Ackermann, No{\"e}mi Aepli,
  Hamid Aghaei, {\v Z}eljko Agi{\'c}, Amir Ahmadi, Lars Ahrenberg,
  Chika~Kennedy Ajede, Salih~Furkan Akkurt, Gabriel{\.e} Aleksandravi{\v
  c}i{\=u}t{\.e}, Ika Alfina, Avner Algom, Khalid Alnajjar, Chiara Alzetta,
  Erik Andersen, Lene Antonsen, Tatsuya Aoyama, Katya Aplonova, Angelina
  Aquino, Carolina Aragon, Glyd Aranes, Maria~Jesus Aranzabe, and et~al. 2023.
\newblock \href {http://hdl.handle.net/11234/1-5287} {Universal dependencies
  2.13}.
\newblock {LINDAT}/{CLARIAH}-{CZ} digital library at the Institute of Formal
  and Applied Linguistics ({{\'U}FAL}), Faculty of Mathematics and Physics,
  Charles University.

\bibitem[{Zhang et~al.(2024)Zhang, Choi, Song, He, Wang, and
  Li}]{Zhang2024HireAL}
Kexun Zhang, Yee~Man Choi, Zhenqiao Song, Taiqi He, William~Yang Wang, and Lei
  Li. 2024.
\newblock \href {https://api.semanticscholar.org/CorpusID:268041426} {Hire a
  linguist!: Learning endangered languages with in-context linguistic
  descriptions}.
\newblock \emph{ArXiv}, abs/2402.18025.

\end{thebibliography}

\appendix

\begin{table*}[t]
\begin{center}
\resizebox{\textwidth}{!}{%
\begin{tabular}{lp{145mm}}
\toprule
\textbf{}
& \textbf{Prompt} \\
\midrule
\textbf{Task Description} &  Please provide the POS tags for each word in the input sentence. The input will be a list of words in the sentence. The output format should be a list of tuples, where each tuple consists of a word from the input text and its corresponding POS tag label from the tag label set: ["ADJ", "ADP", "ADV", "AUX","CCONJ", "DET", "INTJ", "NOUN", "NUM","PART", "PRON", "PROPN", "PUNCT","SCONJ", "SYM", "VERB", "X"].
\\
\textbf{Note}  & Your response should include only a list of tuples, in the order that the words appear in the input sentence, with each tuple containing the corresponding POS tag label for a word.
\\
\textbf{Input}  & ["What", "if", "Google", "Morphed", "Into", "GoogleOS", "?"]  \\

\midrule
\textbf{Output} &  [("What", "PRON"), ("if", "SCONJ"), ("Google", "PROPN"), ("Morphed", "VERB"), ("Into", "ADP"), ("GoogleOS", "PROPN"), ("?", "PUNCT")] \\
\bottomrule
\end{tabular}
}
\caption{\textbf{Prompt template used for POS tagging} based on \citet{lai-etal-2023-chatgpt}. An example prediction by GPT-4}
\label{tab:prompt_templates}
\end{center}
\end{table*}

\section{Models}
\label{appendix_model}
For the experiments, we consider three approaches that are popular in the zero-shot setting since we lack training data for the Brazilian languages. 

\paragraph{Prompting GPT-4} GPT-4~\footnote{\url{https://chat.openai.com/}} is a large language model developed by pre-training on a large amount of texts and code from the web, followed by instruction prompt tuning based on human feedback. We prompt GPT-4 using a similar prompt provided by \citet{lai-etal-2023-chatgpt} where the model is provided a task description before the input. We provide the details in \autoref{appendix_prompt}. 

\paragraph{Cross-lingual transfer} We trained a POS tagger individually for English and Portuguese, and perform zero-shot transfer on other languages. We make use of the XLM-R-large (or simply, XLM-R)~\citep{conneau-etal-2020-unsupervised} for training the models. XLM-R has been pre-trained on 100 languages of the world with over 2TB pre-training corpus size but this corpus does not include any indigenous Brazilian languages. 

\paragraph{Language Adaptive Fine-tuning (LAFT)} We leverage LAFT for an effective cross-lingual transfer by first adapting XLM-R-large model to a new language with limited amount of monolingual data~\citep{alabi-etal-2020-massive,pfeiffer-etal-2020-mad,chau-smith-2021-specializing}. This method was proven to be very effective for low-resource languages~\citep{adelani-etal-2021-masakhaner,muller-etal-2021-unseen}. We make use of the Bible data as the fine-tuning corpus since it is the largest we found for these languages. We only found 7 (out of 12 Brazilian languages) languages with the Bible corpus. Similar to \citet{ebrahimi-kann-2021-adapt}, we examine the effectiveness of this small pre-training corpus with 8K-34K sentences. \citet{pfeiffer-etal-2020-mad} showed that this approach can significantly boost cross-lingual transfer. However, it is not parameter-efficient like the MAD-X they proposed. On the other hand, \citet{ebrahimi-kann-2021-adapt} argued that simple adaptation to a new language is more effective than MAD-X especially when using the Bible corpus for adaptation and we follow this recommendation in our evaluation. 

\paragraph{Hyper-parameter of experiments} For the cross-lingual and LAFT experiments, we used HuggingFace transformers~\citep{wolf-etal-2020-transformers} and A100 Nvidia GPU for fine-tuning the models.  %while for GPT-4 we use the Open AI API. 
For the LAFT, we train for 3 epochs on one GPU while for cross-lingual, we fine-tune English and Portuguese individually using a batch size of 64, with gradient accumulation of 2, and a training epoch of 10. 

\section{Prompt Template}
\label{appendix_prompt}
\autoref{tab:prompt_templates} provides the prompt template we used for GPT-4 evaluation.

%\label{sec:appendix}

%This is an appendix.

\end{document}